\documentclass[11pt]{article}
\usepackage{acl2014}
\usepackage{times}
\usepackage{url}
\usepackage{latexsym}
\usepackage{tikz}
\usepackage{graphicx}

\title{A Comparison of Techniques for Sentiment Classification of Film Reviews}

\author{Milan Gritta \\
  Computer Laboratory \\
  William Gates Building\\
  University of Cambridge \\
  {\tt mg711@cam.ac.uk}}

\date{\today}

\begin{document}
\maketitle
\begin{abstract}
  We undertake the task of comparing lexicon-based sentiment classification of film reviews with machine learning approaches. We look at existing methodologies and attempt to emulate and improve on them using a 'given' lexicon and a bag-of-words approach. We also utilise syntactical information such as part-of-speech and dependency relations. We will show that a simple lexicon-based classification achieves good results however machine learning techniques prove to be the superior tool. We also show that more features does not necessarily deliver better performance as well as elaborate on three further enhancements not tested in this article.
\end{abstract}

\section{Introduction}
Sentiment analysis is a task that utilises natural language processing and text analysis in order to extract the subjective opinion of the author with regards to the topic, review or other entity description. The ultimate goal of this article is to determine the polarity of the sentiment (positive or negative) so that we can harness that information for a specific purpose. There are four main categories of sentiment  analysis. Keyword spotting such as lexicon-based methods looking for presence of words in documents, which we use also use in our paper. \\

Lexical affinity is a more advanced technique that tries to identify the emotional state of the writer/review with higher precision such as 'happy', 'angry' or 'sad'. Statistical methods use machine learning approaches some of which we also use in our paper as a more sophisticated alternative. Concept-level approaches utilise knowledge representation to detect semantics that are too subtle to detect by other conventional means as they are not expressed directly and may require the use of inference.\\

With the recent rise of social media, sentiment analysis has gained a lot in popularity. Big brands are interested in consumer opinion, e-commerce providers want insight into how their products are perceived by their customers, social scientists conduct research into human behaviour online, etc. The use of sentiment classification algorithms allows the processing of huge datasets, which would otherwise be impractical and expensive to process manually. Our goal in this article is to show how simple analysis compares to more sophisticated approaches.

\section{Related Work}
Some of our work resembles \cite{Bo} and we replicate a small subset of that work in our experiments. In their paper, human-produced sentiment analysis had been compared with machine learning approaches. The data used in their experiments comprised 700 positive and 700 negative (IMDb archive) film reviews. The evaluation was performed using cross-validation with three equally sized folds. They used three machine learning algorithms in their experiments. These were Naive Bayes, Maximum Entropy and Support Vector Machines. The results demonstrated that standard machine learning classification outperformed the human baselines of 58\% and 64\% (69\% when humans were allowed to examine the test data) to achieve between 72.8\% and 82.9\% depending on the experiment and the algorithm.

\section{Methods}
 The data we use for our experiments consists of 2000 English film reviews, half of which are positive and half are negative. The reviews contain on average approximately 50 sentences, 900 words or 4000 characters. Next we explain the methodology of our experiments.

\subsection{Lexicon-based classification}
The 'given' lexicon contains 8223 terms formatted in the following manner: weight (weak, strong), length (always 1), target word, part-of-speech (adj, verb, noun, anypos), whether the word is stemmed (yes/no), polarity (positive, negative). The lexicon features allow for at least two comparable approaches, classification with and without weights. We will investigate the impact of using the magnitude (weak, strong) on the resulting classification in order to highlight possible improvements to the system. The lexicon-based method is a purely mechanical string matching process, we later contrast it with more advanced techniques.

\subsubsection*{Negation}
We discussed lexicon-based classification as simple matching of words in the lexicon with the words in film reviews. Based on the scores associated with each word, we add up the total score for each review and decide which side of the polarity threshold it's on. There is another simple technique we investigate in this article, which is using negation in the form of "n't" to statistically evaluate the efficacy of this method. Adding "n't" to the lexicon as a negative sentiment-bearing term may prove a simple, but effective improvement of the system.

\subsection{Bag of words}
Once we established a lexicon-based analysis as a baseline, we will experiment with more advanced machine learning techniques. We conduct different experiments using the bag-of-words machine learning approach. A bag-of-words model is a simple way to represent each document, a film review in our case, with a count of each of the words contained in a document. A bag-of-words is a vector of words as features, which had been assembled by building a dictionary of all words contained in the 2000 reviews, the model uses that vector to describe each film review counting the number of occurrences for each feature. \\

An example vector for a film review may look like this [15, 0, 0, 1, 14, 60, 0, 0, 25, 2, 0, 0, ..., 5]. These vectors are then used to train a classifier. We have found that presence of features as a binary feature works better than feature counts, which was also the case in \cite{Bo}. What this means for the example in the aforementioned explanation is that the final feature vector looks like this [1, 0, 0, 1, 1, 1, 0, 0, 1, 1, 0, 0, ..., 1] and we use this method in all machine learning experiments.

\subsubsection*{Unigrams, Bigrams and Trigrams}
The first experiment uses the simple bag-of-words approach with unigrams (single words) only. We evaluate the quality of analysis based on looking at the presence of individual words in each review. This is called a unigram model, which was one of the methods used by \cite{Bo} and we will make a reference to those results. We then employ combinations of bigrams (word pairs) and trigrams (word triplets) as features to investigate possible improvements. 

\subsubsection*{Words and part-of-speech}
Part of speech (POS) describes the syntactical or morphological behaviour of words. The most commonly known POS categories are nouns, adjectives and verbs and these are the ones we'll be relying on most while not excluding others. The reason POS is helpful in the context of sentiment analysis is that it helps disambiguate some word senses. Knowing that "utter" is a verb allows one to avoid treating it as sentiment amplifier. It's also useful for including or excluding certain categories of words. \\

A simple example of a sentiment-bearing phrase most likely consists of an adjective followed by a noun such as "horrible film". The first POS experiment we conduct is by using words+POS pairs as features for example "great+ADJ" or "film+NOUN". We then augment this approach by using words plus word+POS for instance "great great+ADJ" or "film film+NOUN" and perform statistical analysis of significance. The final POS method is using only particular POS to train the classifier. We run three such experiments with adjectives, nouns and verbs only.

\subsubsection*{Lexicon as features}
We proceed to use the sentiment lexicon features to enhance the standard bag-of-words in order to investigate whether an augmentation using a fixed lexicon-based feature set would improve the baseline model. The methodology is explained by enumerating all 6887 unique words from the lexicon in a feature vector and merging it with the bag-of-words obtained from the reviews. We then train a classifier on this enhanced feature set, which is the reason this approach might do better than baseline.

\subsubsection*{Dependency features}
We use Dependency Parsing (DP) in this experiment to enhance the features and improve on previous techniques. DP is based on the theory of Dependency Grammar where the verb is treated as the structural centre of the phrase/clause/sentence. All other words in the sentence are directly or indirectly dependent on the verb. DP outputs a tree which describes the relations and dependencies between individual words. An example tree is depicted below:

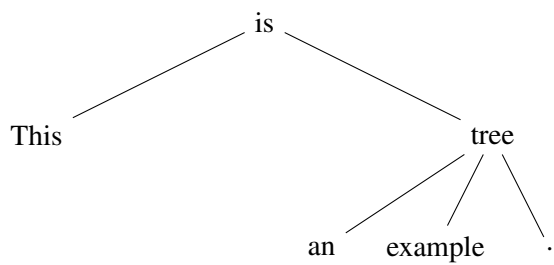
\begin{center}
\begin{tikzpicture}
	\node (is-root) {is}
		[sibling distance=6cm]
		child { node {This} }
		child {
			node {tree}
				[sibling distance=1.5cm]
				child { node {an} }
				child { node {example} }
				child { node {.} }
				child[missing]
		};
	\path (is-root) +(0,-2.5\tikzleveldistance)
		node {Parsed: \textit{This is an example tree.}};
\end{tikzpicture}
\end{center}

We choose to utilise the \textit{typed} dependencies, which look like this: nsubj(tree-5, This-1), cop(tree-5, is-2), det(tree-5, an-3), nn(tree-5, example-4), root(ROOT-0, tree-5) for our experiments. The numbers above denote the position of text in the sentence, which we do not use as \cite{Bo} showed that it does not improve classification accuracy. \\

We use the labels ("cop", "nsubj", "amod", ...) as additional syntactical knowledge to pair a word with it's dependency label to obtain a new feature pair. This may take form as the following simple feature vector: ["This+nsubj", "is+cop", "an+det", "example+nn", ..., ..., etc.]. A pair may be further augmented by appending part-of-speech to it or some other meaningful information.

\subsection{Algorithms}

\subsubsection*{Naive Bayes}
The Naive Bayes (NB) is a probabilistic classifier based on applying the Bayes' Theorem, which relates current belief to prior belief. It can be seen as a way of understanding how the probability that a theory is true is affected by a new piece of evidence. It's described in the following simple and elegant equation:

\[ P(h|D) = \frac{P(D|h) P(h)}{P(D)} \]
\begin{center}
{\small Bayes' Theorem}
\end{center}
\begin{itemize}
\item $P(h)$ = prior probability of hypothesis $h$
\item $P(D)$ = prior probability of training data $D$
\item $P(h|D)$ = probability of $h$ given $D$
\item $P(D|h)$ = probability of $D$ given $h$
\end{itemize}

It's particularly popular in text categorisation, which is why we use it in our experiments. It is also very scalable thus a great choice for a bag-of-words model running into 000's of dimensions. The scalability and simplicity of NB comes from its assumption of the independence of features, which is also the reason for the 'Naive' in NB. In spite of its naivety, the model "is surprisingly effective in practice since its classification decision may often be correct even if its probability estimates are inaccurate" \cite{RI}. \\

Bayesian classifiers assign the most likely class to a given example described by its vector. Naive Bayes works best in two cases: completely independent features (as expected) and functionally dependent features (which is less obvious), while reaching its worst performance between these extremes \cite{RI}. This is promising since the bag-of-words approach contains words, phrases, sentences and general discourse of highly interdependent features, which leads us to expect good performance from the NB classifier.

\subsubsection*{Support Vector Machines}
A support vector machine (SVM) is a supervised learning model used for regression and classification. It handles data with only two classes such as positive or negative as is the case with the film review data. This algorithm can also be applied to non-linearly separable binary data. SVM represents data as points in high-dimensional space. It then proceeds to separate the two classes by a maximum margin hyperplane (a vector maximising the gap between classes). Classification is then a simple process of dividing the samples into two classes based on which side of the hyperplane they fall on. \\

More technically, we have \textit{L} training points, where each input $x_{i}$ has \textit{D} attributes (i.e. is of dimensionality \textit{D}) and is in one of two classes $y_{i}$ = 1 or 0, i.e our training data is of the form:

\[\{x_{i}, y_{i}\}\ where\ i = 1\ ...\ L, y_{i} \in \{1,0\},x \in R^{D}\]
The hyperplane is described by $w \cdot x + b = 0$ where \textit{w} is normal to the hyperplane and $\frac{b}{\left\vert\left\vert{w}\right\vert\right\vert}$ is the perpendicular distance from the hyperplane to the origin.\\

We use the machine learning toolkit WEKA \cite{WEKA}, which contains a variant of SVM called Sequential Minimal Optimisation. It's is an implementation of John C. Platt's sequential minimal optimization algorithm for training a support vector machine classifier \cite{JP}. SVM algorithm is very robust with respect to overfitting since the boundary separating the two classes is decided on by looking at just a few key instances rather than the entire dataset. This is why we chose it as one of three algorithms for this task.

\subsection{Maximum Entropy}
A Maximum Entropy classifier (ME) is a probabilistic classification method widely used in text processing, which generalises logistic regression to problems with more than two discrete outcomes. ME is closely related to the Naive Bayes classifier however rather than treating each feature as independent, the model finds weights for features that maximise the likelihood of the training data. It is especially useful when knowledge about the prior distributions is absent and we cannot make any assumptions about feature independence. \\

Because of this difference, learning a ME model takes longer than the NB model. Whereas with NB we only have to count co-occurrences of features and classes, with ME we maximise the weights using an iterative procedure. The probability \textit{p} of class \textit{c} given document \textit{d} is given by the following equation where the $\frac{1}{Z(d)}$ is a normalisation factor:

\begin{equation}
p(c | d)=\frac{1}{Z(d)} \exp \left(\sum_{i} \lambda_{i} f_{i}(d, c)\right)
\end{equation}

The $\lambda _{i}$ is the $i^{th}$ feature-weight parameter where a large $\lambda _{i}$ means that the feature $f_{i}(d, c)$ is considered a strong indicator for class \textit{c}. ME is founded on the principle of maximum entropy meaning that without external knowledge, one should prefer distributions that are uniform i.e. have maximal entropy. ME estimates the conditional distribution of the class label given a document, which is represented by a set of features. \\

The labelled training data is used to estimate the expected value of these features on a class by class basis. For instance, in a 3-way classification task where we know that 50\% of the documents containing "student" belong to the 'university' class, we can say that a  document with "student" in it has a 50\% chance of being a 'university' document and a 25\% chance of coming from the other two classes. If a document doesn't have "student" in it, we uniformly guess a 1/3 chance for each of the classes.

\section{Results}
We use a 3-fold cross-validation (CV) in our experiments, which is a technique for model validation in statistical analysis. CV aims to avoid overfitting in training a classifier. It does this by randomly partitioning the data into equally sized folds (or sets), in our case three and proceeds to test the classifier on each of the folds while training on the remaining two. This procedure is also referred to as the "repeated holdout". To ensure the accuracy of the comparison of all systems, we fix the random seed of the WEKA cross-validation function in order to enforce the constancy of folds across statistical significance tests. \\

For hypothesis testing, we use the Wilcoxon signed-rank test. This is a statistical hypothesis test comparing two related samples for repeated measurements to assess whether their population mean ranks differ. In our case it's the baseline comparison with an alternative approach, for instance using the lexicon \emph{with} and \emph{without} weights to determine significance levels. We use it as an alternative to the Student's t-test since we cannot assume that the samples are normally distributed. We report significance in all of the following experiments. As the review data are balanced, our evaluation metric is as simple classification accuracy expressed as a percentage.

\subsection*{More features, same precision}
During the course of our research we noticed that the number of features used for training the classifiers is robust to change. We ran a short experiment to confirm our hypothesis that varying the number of features from 5000 to 10000 to 20000 does not change the resulting accuracy. The  test allowed us to reject the null hypothesis that classification results will change with 99.5\% confidence level. Also the accuracy remained unchanged in all three experiments at 79.59\%. We therefore decided to use 10000 features in all machine learning experiments with the exception of using the Maximum Entropy classifier, which requires more memory due to its inherent complexity. For those experiments only, we used 5000 features.

\subsection{Lexicon-based classification}
We established the baseline at 64\% correctly classified instances using only words where a matching positive word adds 1 point while a negative match subtracts 1 point from the overall score. We then enhanced this basic method with weights so that a strong positive match now adds 5 points and a strong negative match takes 5 away in order to markedly differentiate individual magnitudes. A weak positive or weak negative match counts as 1 point as with the basic method. \\

The new model (with magnitude) achieved a 2.5\% increase in accuracy. One additional augmentation was adding the negation string "n't" to the lexicon as a strong negative term. This simple step rendered the performance to peek at 69.7\%, which was impressive for such a simple model.

\subsection{Unigrams, Bigrams, Trigrams}
From now onwards, we will be presenting the results of machine learning approaches to contrast with the lexicon-based methods. To establish a baseline, we ran the bag-of-words experiment with unigrams only. This gave a Naive Bayes (NB) baseline of 79.59\%, Maximum Entropy (ME) baseline of 77.21\% and a Support Vector Machine (SVM) baseline of 85.03\%. We also conducted a significance tests between NB, ME and SVM pairs to check that the various approaches are statistically different so that we were justified in using these algorithms in a comparative manner.\\

We obtained 79.39\% (SVM), 77.52\% (ME) and 75.91\% (NB) for bigrams, which is similar to the $\sim$77\% in \cite{Bo}. With trigrams, accuracy slightly decreased except for ME achieving 84.47\%. The best system out of all experiments for the SVM classifier peaked at 86.59\%, which used features comprising unigrams and bigrams simultaneously. This was almost 4\% better than \cite{Bo} however the accuracy for NB was almost identical while ME got 84.22\%. When we combined all three, that is unigrams, bigrams and trigrams into one feature vector, the NB classifer obtained its best result of all topping 80.54\%, which was only a fraction better than using unigrams and bigrams. The addition of trigrams actually made the SVM dip by 0.4\% while ME results were virtually unchanged. That suggests to us that trigrams add little or no value to a feature set, something we observed with all classifiers.

\subsection{Words and part-of-speech}
When we added part-of-speech (POS) to unigrams, it boosted NB by just under 1\%, ME by over 2\% and decreased SVM accuracy by just over 1\%. Later using this combination of features [word word+POS, word2 word2+POS2, ...], there was a significant loss of precision (6\%) for ME while for NB and SVM there was no improvement on the baseline despite our positive expectations. We then ran experiments with individual pos tags only to see how much a model can be simplified before losing its effectiveness and how much does the performance degrade if at all. \\

For adjectives, we obtained a respectable 76.11\% (SVM), 73.48\% (ME) and 77.52\% (NB), which was almost identical to \cite{Bo}. For nouns, there was a noticeable drop to 75.55\% (SVM), 70.76\% (ME) and 70.41\% (NB). However the most degradation we saw was using only verbs obtaining just 66.69\% (SVM), 63.45\% (ME) and 67.01\% (NB). This experiment indicates that perhaps unsurprisingly, adjectives are the most expressive part of speech in film reviews. Although the peak accuracy is some 10\% off the best system, we have highlighted that even the simplest methods such as only adjectives as features still outperform the lexicon-based system by $\sim$7\%. Combining all three pos tags did not produce statistically significant results for any of our machine learning algorithms.

\subsection{Lexicon as features}
The lexicon-enhanced bag-of-words machine learning system achieved 84.74\% (SVM) and 79.25\% (NB), which is less than 0.5\% difference compared to baseline. This is because the "new" feature vector is less than 0.5\% longer than the standard. Let us elaborate: The lexicon contains 6887 common sentiment bearing words. The 2000 reviews contain $\sim$1.8M words, which is a large enough corpus to capture almost all 6887 lexicon words. \\

The hypothesis that using a lexicon to enhance the feature set leads to improved classification would have to be tested on a smaller set of reviews where the corpus, from which the bag-of-words is built, is much smaller. In this manner, the lexicon words might provide enough additional features to improve overall performance. In our experiment, the lexicon does not contain enough new distinguishing information to deliver a performance boost. ME classifier results did not meet the 95\% confidence interval requirement.

\subsection{Dependency parsing}
Possessing the representation of the grammatical relations between words in a sentence can be exploited as an added feature in our enhanced bag-of-words approach. When we augmented the system with words as features + textual relation (for instance "This+nsubj"), we obtained 81.85\% (SVM) and 78.13\% (NB), which is $\sim$2\% lower than using POS. We think the failure to improve the model despite possessing high quality dependency relations lies in our application rather than the usefulness of those features. \\

We think a better approach would be extracting verb and noun phrase chunks as bigrams and trigrams from a parse tree such as can be seen below. Indeed this would be suitable experiment to conduct in further work. Further research led us to \cite{FOREST}, which uses Dependency Forests (DF) to eliminate one major drawback of the 1-best dependency tree feature extraction, the parsing errors. In a 1-best dependency tree, several parsing alternatives are considered before the best annotation is chosen. \\

A DF is an encoding of multiple dependency trees in a compact representation, which provides an elegant solution to the problem of parsing error propagation. They cite 7\% and 12\% parsing errors for Engligh and Chinese respectively, which may have accounted for the drop in classification accuracy in our experiments. Parsing errors caused by irregular language used in online reviews make DF an attractive augmentation of a sentiment classification system. Employing DF in their experiments, \cite{FOREST} achieved an impressive best score of 91.6\% accuracy.\\

An approach of this kind captures more intelligent features than just generating arbitrary bigrams and trigrams. It selectively constructs key phrases based on the most information bearing terms thus theoretically improving the basic model. Unexpectedly, the ME classifier achieved its best result of 84.67\% with very high statistical significance, which is more than 7\% better than baseline while the other two algorithms struggled.

\subsection{Multi Classifier Systems}
In the final stages of our research, we came upon a serendipitous discovery. Having performed over 40 different experiments, we tested an ad hoc hypothesis that using the 40+ recorded predictions for each instance, we could calculate the mode as a classification decision. Deploying this "Combo Classifer" (Ensemble Learning) we were able to correctly predict 85.1\% of the instances with a very high statistical significance. \\

This closely resembles the Multi Classifier Systems (MCS) methodology reviewed by \cite{OX}. MCS fuses together multiple outputs from weak classifiers for better accuracy and classification. While we have not set out to implement this approach, it was fascinating to see the emergent methodology as a possible classification mechanism. Its accuracy was only 1.5\% lower than the best system tested in the paper.

\begin{table}[h]
\begin{center}
\begin{tabular}{l|r|r}
\hline \bf Features & \bf \% Correct & \bf Significant \\ \hline
no weights & 64.0 & baseline\\
with weights & 66.5 & Yes\\
weights + negation & 69.7 & Yes\\
"Combo Classifier" & 85.1 & Yes\\
\hline
\end{tabular}
\end{center}
\caption{Lexicon and Multi Classifier Systems}
\end{table}

\begin{table}[h]
\begin{center}
\begin{tabular}{l|rrrr|}
\hline \bf Features & \bf SVM & \bf NB & \bf ME\\ \hline
unigrams (uni) & *85.03 & *79.59 & *77.21\\
bigrams & \textbf{79.39} & \textbf{75.91} & \textbf{77.52}\\
trigrams & \textbf{79.89} & \textbf{74.25} & \textbf{84.47}\\
1, 2 grams & \textbf{86.59} & \textbf{80.14} & \textbf{84.22}\\
1, 2, 3 grams & \textbf{86.19} & \textbf{80.54} & \textbf{84.38}\\
uni+pos & \textbf{83.77} & \textbf{80.39} & \textbf{81.85}\\
uni+uni+pos & \textbf{84.57} & 80.85 & \textbf{75.95}\\
lexicon added & \textbf{84.74} & \textbf{79.25} & 83.48\\
adjectives & \textbf{76.11} & \textbf{77.52} & \textbf{73.48}\\
verbs & \textbf{66.69} & \textbf{67.01} & \textbf{63.45}\\
nouns & \textbf{75.55} & \textbf{70.41} & \textbf{70.76}\\
adj+verb+noun & 83.47 & 78.78 & 80.24\\
dependencies & \textbf{81.85} & \textbf{78.13} & \textbf{84.67}\\
\hline
\end{tabular}
\end{center}
\caption{
Machine Learning Accuracy in percent, statistically significant results in bold face
    \begin{tabular}{l l}
      \textbf{SVM} & Support Vector Machine Algorithm\\
      \textbf{ME} & Maximum Entropy Classifer\\
      \textbf{NB} & Naive Bayes Classifier\\
      \textbf{*} & Baseline / Benchmark\\
    \end{tabular}
}
\end{table}

\section{Future Work}

We now propose further alternatives for the enhancement of the features used for sentiment analysis. We particularly favour three possible future augmentations to our system. These are:  Utilising lexical information from WordNet, performing anaphora resolution and modelling discourse effects in reviews. We suggest how one may approach these individually.

\subsection{Utilising WordNet}
The first method of feature augmentation is utilising WordNet (WN). WN is a large lexical database of English maintained by Princeton University. Nouns, verbs, adjectives and adverbs are grouped into sets of cognitive synonyms (synsets), each expressing a distinct concept. We can use the synsets, which express a particular word sense, to represent words with their senses as features. \\

\cite{WN} used this approach successfully to improve on the baseline (unigram bag-of-words) of 84.9\% to achieve a modest gain of just over 1\% for automatic sense induction and a gain of just over 5\% for manually annotated word senses. In any natural language processing task, we try to minimise or eliminate any manual annotation as it is expensive and time-consuming, so feature augmentation using WN a possible improvement to our system if only with modest gain.

\subsection{Anaphora Resolution}
In linguistics, anaphora refers to an expression, which in order to be interpreted, depends on some other expression in context. For example in the sentence "The president chose to veto the latest bill, but \textbf{he} wasn't 100\% sure.", \textbf{he} is an \textit{anaphoric} pronoun, which refers to the president. In the context of film reviews reviews, for example in a sentence "..., \textbf{it} was a terrible experience.", it's useful to determine what \textbf{it} is referring to. This allows us to judge the relevance of that statement in the context of the review. \\

\cite{AN} experimented with anaphora resolution (AS) to improve opinion mining (OM) in film reviews. They found that adding AR to OM significantly boosted recall in target extraction although the precision decreased leading to a slightly lower F-measure. Their article also presents extensions to address this problem and so it may be worthwhile to include AS in our system as well. 

\subsection{Discourse Analysis}
Sentiment analysis can be further enhanced using Discourse Analysis (DA). The aim of DA is to analyse linguistic content such as a conversation looking beyond the sentence boundary. Rather than focusing only at the word level of what is being said, DA takes into consideration the surrounding contexts at the sentence level. It helps to discover whether each sentence is central to the overall point by identifying connectors \textit{(words and phrases)} that signal a shift or continuation in the discourse structure. \\

\cite{DIS} showed a significant improvement in sentiment analysis scores employing DA. They recorded a 91.36\% accuracy using connectors compared to the baseline (unigram bag-of-words) of 84.79\% using the standard SVM. We suggest using their approach in future work to boost the overall system performance, a sentiment which is mirrored by \cite{Bo} in their discussion section on error analysis and the style of rhetoric encountered in reviews.

\section{Conclusion}
We conducted comparative experiments in the field of sentiment analysis of film reviews contrasting lexicon-based techniques with more advanced machine learning approaches. We highlighted related experiments and their individual performance gains, however it remains unknown how these augmentations impact on performance once combined into one classification model. The closest we got to a combination classifier was the MCS system achieving 85.1\% by pooling together the scores from other weaker classifiers. We also showed that more features does not necessarily translate to better performance, an observation made and tested during the course our our research. \\

In the case of dependency relation features, we suggested extracting phrasal chunks from parse trees rather than appending the typed dependency labels to the word as a better alternative after obtaining unimpressive results in that particular experiment. Related work by \cite{FOREST} makes a suitable candidate for further improvement by using Dependency Forests eliminating parsing errors. Just as in \cite{Bo}, Support Vector Machine algorithm proved to be superior to Maximum Entropy, which in turn was superior to the Naive Bayes algorithm. \\

As we explained earlier, unlike Naive Bayes, Maximum Entropy makes no assumptions about the relationships between features, and so it performed better when conditional independence assumptions were not met. The main conclusion to take away is that the simple machine learning system of unigrams as bag-of-words works incredibly well. We engineered many augmentations and performed some 40 experiments, yet the fact remains that the best performing classification system was a mere $\sim$1\% better than the simplest one (unigrams).

\end{document}